\documentclass[conference]{IEEEtran}
\IEEEoverridecommandlockouts
\usepackage{cite}
\usepackage{amsmath,amssymb,amsfonts}
\usepackage[ruled,vlined]{algorithm2e}
\usepackage{graphicx}
\usepackage{textcomp}
\usepackage{xcolor}
\usepackage{lmodern}
\usepackage{subcaption}
\usepackage{amsmath}
\usepackage{bbm}
\usepackage{hyperref}
\def\BibTeX{{\rm B\kern-.05em{\sc i\kern-.025em b}\kern-.08em
    T\kern-.1667em\lower.7ex\hbox{E}\kern-.125emX}}
\begin{document}

\title{Using a Local Surrogate Model to Interpret Temporal Shifts in Global Annual Data}
\author{\IEEEauthorblockN{Shou Nakano}
\IEEEauthorblockA{
\textit{Wilfrid Laurier Univeristy}\\
Waterloo, Canada \\
naka2910@mylaurier.ca}
\and
\IEEEauthorblockN{Yang Liu}
\IEEEauthorblockA{
\textit{Wilfrid Laurier University}\\
Waterloo, Canada \\
yangliu@wlu.ca}
}

\maketitle

\begin{abstract}
This paper focuses on explaining changes over time in globally-sourced, annual temporal data, with the specific objective of identifying pivotal factors that contribute to these temporal shifts. Leveraging such analytical frameworks can yield transformative impacts, including the informed refinement of public policy and the identification of key drivers affecting a country's economic evolution. We employ Local Interpretable Model-agnostic Explanations (LIME) \cite{b3} to shed light on national happiness indices, economic freedom, and population metrics, spanning variable time frames. Acknowledging the presence of missing values, we employ three imputation approaches to generate robust multivariate time-series datasets apt for LIME's input requirements. Our methodology's efficacy is substantiated through a series of empirical evaluations involving multiple datasets. These evaluations include comparative analyses against random feature selection, correlation with real-world events as elucidated by LIME, and validation through Individual Conditional Expectation (ICE) \cite{b17} plots, a state-of-the-art technique proficient in feature importance detection.
\end{abstract}


\section{Introduction}

This paper investigates temporal data collected from global sources on an annual basis. Our primary focus lies in the effective utilization of advanced data analytics to forecast emerging trends and identify avenues for improvements at the national level. The insights gained from these analyses have far-reaching implications, including the potential to guide policy reforms that could significantly enhance the well-being of citizens. Additionally, our work offers valuable perspectives on the key factors that play a pivotal role in shaping a country's economic direction, among other applications.

LIME has traditionally been employed to dissect and interpret model predictions. Its adaptability has been showcased in areas like time-series data analysis \cite{b1} and the interpretation of cardiographic signals \cite{b2}. Furthermore, it has been pivotal in formulating explanations that precisely illuminate a model's workings. A significant innovation in this study is the extension of LIME's scope to encompass datasets featuring multiple time-series, marking a departure from the customary singular time-series modality. This paradigm shift is paramount for dissecting annual changes across nations. To facilitate this, we propose a list of approaches to handle missing values and reform datasets of different formats. 

In addition, LIME is deployed to scrutinize predictions generated by models fine-tuned using FLAML \cite{b13}, a Python library adept at autonomously pinpointing optimal hyperparameters across a spectrum of intricate models. By delving into an assortment of black-box models, this research aims to elucidate LIME's proficiency in distilling these models and elucidating the foundational rationale of their predictions. Furthermore, it offers insights into how alterations in features influence the predicted outcomes, enhancing user comprehension. 

This study employs three open-access time-series datasets that capture variations in national happiness levels, economic freedom, and population dynamics. We assess the validity of our approach through a series of tests that include: benchmarking against randomly selected features, establishing correlations with real-world occurrences as interpreted by LIME, and cross-validating outcomes using Individual Conditional Expectation (ICE) plots, a leading method for determining feature significance. The focus lies on the precision of LIME in feature identification, supported by concrete case studies that authenticate its predictive claims, regardless of whether they carry positive or negative implications.

\section{Related Work}

Introduced in 2016, LIME aimed to understand black box models by generating interpretable and accurate predictions via the creation of a more transparent model \cite{b3}. This transparency is achieved by iteratively sampling the intricate model to approximate its behavior. Given LIME's approach of treating the model as a black-box, it exhibits model-agnostic properties. Then, from LIME's recreation of the model, individual points can be provided for LIME to provide explanations for. This approach was expanded upon by several researchers who noted that LIME lacked any support for time-series data with thousands of entries, such as hourly weather updates or stock prices, leading to the proposal of LEFTIST, the first model-agnostic approach for time series data\cite{b1}, which segmented the time series data into several components of equal length. This was followed by TS-MULE \cite{b4}, which approached several problems like how to properly segment the time series into several windows of different lengths, such as SAX segmentation and suggested several transformations. Finally, LIMESegment was developed a year later \cite{b5}, which outperformed previous research with an approach involving frequency distributions and understanding the local neighborhood of a time series.

Previous work has also been done in terms of applying LIME to various datasets, since LIME is highly successful and can be easily applied to textual, tabular and image data. For example, LIME has been used to categorize species of plants after InceptionV3, XCeption and ResNet50 models were trained \cite{b6}, to show why image captioning models selected certain words to be a part of an image’s caption \cite{b7}, to classify healthy and unhealthy hair through pictures of various scalps in order to detect diseases like folliculitis decalvans and acne keloidalis, with a high accuracy of 96.63\% from the black box model that was backed up with explainable machine learning.

Several other applications exist, such as using LIME in conjunction with other approaches like Shapley Values to explain why users could potentially have diseases like diabetes and breast cancer \cite{b8}. LIME has also been used to provide real-world explanations for why a synthetic dataset of various machine parts would suffer breakdowns, with explanations being provided based on the parts' wear and tear, as well as the temperatures that they were exposed to \cite{b9}. From the observations, it's evident that LIME can efficiently generate insightful explanations for a model, provided the model is precise and the data type is compatible with LIME or its variants like LIMESegment. However, for the scope of this study, LIMESegment and other approaches to univariate time-series are not valid choices, since LIMESegment works with univariate time series, not multivariate ones. Thus, an algorithm like LIMESegment will only be able to look at the year column and the target value, instead of combining them with other features to glean more information. Therefore, a strategy that encapsulates the entirety of the data and works with any tabular dataset, not just a specific group of datasets is essential, ensuring the model has an optimal amount of data for processing.

\section{Methodology}

\subsection{Problem Definition}

In this study, our objectives are twofold. Firstly, given an input \( X = (x_1, x_2, \ldots, x_n) \), we aim to predict the target label \( y \) with the highest accuracy utilizing sophisticated models. The data at hand might necessitate imputation and is bound to undergo adjustments for temporal scrutiny. The adoption of intricate models serves to demonstrate LIME's capability to render a model more comprehensible irrespective of its inherent complexity. Formally, LIME can be mathematically represented as Equation (1). Here, \( L \) denotes the loss, capturing the discrepancy between LIME's approximated model, in this case, a linear model \( g \) and the actual model \( f \). \( \pi_x \) delineates the extent of the vicinity around \( x' \), the point that was passed into LIME, where the vicinity is where points will be sampled, while \( \Omega(g) \) signifies the model's complexity. Both \( L \) and \( \Omega(g) \) should be minimized.

\begin{equation}
explanation(x)= \underset{g\in G}{\arg\min} L(f,g, \pi_x) + \Omega(g)
\end{equation}

Secondly, our aim is to showcase the precision and capabilities of LIME when applied to the adapted data. This is achieved by comparing LIME's outcomes and predictions with real-world instances, and by contrasting the Individual Conditional Expectation Plots with LIME's adaptation of the Submodular Pick \cite{b17}. The latter addresses an NP-Hard challenge, where the objective is to generate explanations for points, ensuring maximal distinctiveness. The underlying mathematics for the submodular pick algorithm can be shown through Equation (2) and Equation (3).

\begin{equation}
c(V, W, I)= \sum_{j=1}^{d'}  \mathbbm{1}_{[\exists i\in V: W_{i,j}>0]}I_j
\end{equation}

\begin{equation}
Pick(W,I) = \underset{V, |V|\leqslant B}{\arg\max} c(V,W, I)
\end{equation}

Equation (2) offers a definition of coverage by examining \( V \), a collection of instances, alongside \( W \), representing the local significance of each instance. \( I_j \) symbolizes the global prominence of a specific element within the explanation domain. The formulation of \( I_j \) is contingent upon the data type and can be refined for enhanced efficiency. Equation (3) focuses on the pick problem, striving to amplify the cumulative coverage achieved by selecting multiple local instances. This is realized by leveraging the values computed from Equation (2).

\subsection{Data Collection}

Three datasets were collected for the purpose of this experiment, where the first originates from the World Happiness Report \cite{b10}, the second dataset was collected from Kaggle \cite{b11}, which was initially sourced from the Fraser Institute and the third dataset comes from the U.S.A.'s Census Bureau \cite{b12}. Each dataset is predominantly numerical, aside from the names of countries. Additionally, the values in the first two datasets range from 0 to 10.

\noindent {\bf Dataset 1:} The World Happiness Report dataset \cite{b10} compiles yearly data from 2015 to 2022 by gathering polls from thousands of citizens to understand how they feel about their country and these results are then used to predict a happiness score. 
Over the years, the data collection methods, naming conventions, and columns have evolved. The final dataset retained columns such as Country Name, Year, Life Ladder, GDP per Capita, Social Support, Healthy Life Expectancy, Freedom to Make Life Choices, Generosity, Perception of Corruption, and Dystopia + Residual. Country names that had changed from 2015 to 2022, like Swaziland to Eswatini, were manually updated. This dataset was chosen to investigate if any particular features would greatly affect a country's happiness score.

\noindent {\bf Dataset 2:} The second dataset was originally published by the Fraser Institute and later reformatted on Kaggle \cite{b11}, focuses on the economic freedom of various countries. It is organized into five main categories, each containing multiple sub-categories, totaling 25 factors. Each country's economic freedom score lies between 0 and 10. Such categories include military interference, inflation, business regulations, etc. The dataset's linear relationships allow for a relatively easy calculation of a country's economic freedom, yet it also serves as a useful tool for case studies and evaluating the effectiveness of LIME with FLAML-trained models.

\noindent {\bf Dataset 3:} The third dataset is a comprehensive set of data published by the U.S.A. \cite{b12}, which tracks countries' population year by year and provides a detailed background of its' sex ratio, fertility rate, life expectancy, mortality rate, crude death rate, among others. This dataset provides hypothetical estimates of the population up to 2100, but for the sake of realism in this report, the dataset has been truncated at 2023. This dataset aims to evaluate how real-world, non-calculated Y-column data performs in temporal analyses and with the LIME approach.

\subsection{Preprocessing}

\noindent {\bf Addressing Missing Data:}
Managing data with missing values is a pivotal issue in both data analytics and model building. This concern is especially pronounced in our investigation involving the second dataset, where nearly $65\%$ of the entries had at least one column with missing data. To tackle this challenge, we utilized three distinct imputation methods, ultimately selecting the best-performing approach for data imputation. This process is illuatrated with Algorithm \ref{alg:missing}.
\begin{enumerate}
\item Linear Regression Imputation: To predict missing values, we treat the feature with missing data as the dependent variable and use the other features which are complete (or mostly complete) as the independent variables. 
Formally, if a value \( y_i \) is missing, its imputed counterpart \( \hat{y}_i \) can be calculated using the equation:
\[
\hat{y}_i = \hat{\beta}_0 + \hat{\beta}_1 x_{i1} + \hat{\beta}_2 x_{i2} + \ldots + \hat{\beta}_n x_{in}
\]
Here, \( \hat{\beta}_0, \hat{\beta}_1, \ldots, \hat{\beta}_n \) represent the estimated coefficients obtained from the linear regression model trained on the observed data.
The imputed value \( \hat{y}_i \) then takes the place of the missing value. Before using the model for imputation, we adopt Ordinary Least Squares (OLS), which minimizes the sum of squared differences between observed and predicted values, to estimate $\beta$. 
This approach intuitively leverages relationships between variables to estimate absent data, ensuring that the inherent correlations in the dataset are considered. 
\item KNN Imputation: We estimate a missing value by considering its `k' most similar data points (neighbors) that have no missing values for that specific feature. The task is further divided into two steps:
\begin{enumerate}
\item Finding Neighbors:
We identify the \(k\) nearest neighbors based on Euclidean distance metric
\[
d(x, y) = \sqrt{\sum_{i=1}^{n} (x_i - y_i)^2}.
\]
\item Computing the Imputed Value:
The imputed value \(x_{\text{imputed}}\) is then calculated as the weighted average of these \(k\) neighbors:
\[
x_{\text{imputed}} = \frac{\sum_{i=1}^{k} w_i x_i}{\sum_{i=1}^{k} w_i}
\]
where \(w_i\) are the weights, corresponding to the inverse of the distance to the neighbors.
\end{enumerate}
This method offers the advantage of flexibility, as it is suitable for both numerical and categorical data and does not rely on linear assumptions, allowing it to handle non-linear relationships in datasets effectively. However, the method can be computationally intensive, especially for large datasets, due to the need to calculate distances and identify nearest neighbors. 
\item Iterative Imputation: We take a stepwise approach to model each feature with missing values as a function of other features in an iterative manner.
The process begins with an initial imputation with mean values. Each feature with missing values is then treated as a dependent variable, with the remaining features acting as independent variables. The missing values are predicted iteratively, cycling through each feature, until the imputations stabilize.  Mathematically, the imputed value for a missing entry \( x_{ij} \) at iteration \( t \) is given by \( x_{ij}^{(t)} = \beta_0^{(t)} + \beta_1^{(t)} x_{1j} + \ldots + \beta_n^{(t)} x_{nj} + \epsilon_j^{(t)} \), where \( \beta_0^{(t)}, \beta_1^{(t)}, \ldots, \beta_n^{(t)} \) are the estimated coefficients and \( \epsilon_j^{(t)} \) is a random residual.

\end{enumerate}

\begin{algorithm}
\caption{Imputation Algorithm for Missing Data}\label{alg:missing}
\KwData{Two-dimensional dataset \( \texttt{D} \) with missing values, threshold \( \theta \) for missing rate}
\KwResult{Imputed dataset}

\BlankLine
Initialize an empty dataset \( \texttt{imputed\_data} \)\;
\ForEach{\( \text{row in } \texttt{D} \)}{
  \texttt{missing\_count} \( \leftarrow \) find\_missing(row)\;
  \texttt{total\_count} \( \leftarrow \) length(row)\;
  \texttt{missing\_rate} \( \leftarrow \) \texttt{missing\_count} / \texttt{total\_count}\;
  
  \eIf{\texttt{missing\_rate} \( \leq \theta \)}{
    \texttt{imputed\_row}\( \leftarrow \) impute\_model(row)\;
append(\texttt{imputed\_data}, \texttt{imputed\_row});
  }{
    append(\texttt{imputed\_data}, row)\;
  }
}
\Return{\texttt{imputed\_data}}\;

\end{algorithm}

In this work, the linear regression imputation was selected due to its superior performance in retaining the quality of the dataset, in contrast to KNN Imputation and Iterative Imputation which resulted in a decline in the $R^2$  score by $0.08-0.1$ in tests. Additionally, we chose to impute rows which had 8 or less missing columns, which set a threshold for a row, where a row must have at least 75\% of its' possibly missing data available in order to have its' remaining columns imputed. Dataset 2 had 36 columns and 33 of them could have missing values. This allows for a good balance between the amount of data imputed and the quality of data, since around 1,000 rows were saved with this technique.\\
\noindent {\bf Reformatting Data for Analysis:}
Our primary goal here is to discern which features most significantly influenced the shift in a country's performance, rather than focusing on specific yearly values. Given the distinct characteristics of each dataset, tailored strategies were adopted to reformat the original data into temporal multivariate timeseries datasets. For the first and second datasets, apart from the country names, each row's values were deducted by the preceding row's values, provided both rows pertained to the same country. This process enabled the capture of year-to-year changes. On the other hand, the third dataset, due to its unique structure, necessitated a different approach. The only column where differences were computed was the population. For other columns, values from the preceding row were directly used to emphasize relative proportions rather than the absolute variations across columns. It's worth noting that these strategies have the potential to be extended to various applications that possess a similar data format. Random samples are shown in the Appendix to demonstrate the unique nature of different datasets.

\subsection{Building Models with FLAML}

Once this was done, the countries' names were encoded into numerical values, for the purpose of being able to use them with LIME later on. Then, the data was split with a 80-20 split and trained through the use of the Python package FLAML \cite{b13}. FLAML is built for machine learning, particularly for cost-effective hyperparameter optimization for ensemble classification and regression methods. The statistic that should be optimized, such as the model's RMSE or R\textsuperscript{2} value can be manually selected, as well as the amount of time to be given to the algorithm and the algorithm will provide regular updates on its' progress. As seen in Fig 1., the models built by FLAML, whether they're with Light GBM or ExtraTrees provide a near-optimal fit for the data, with the line of best fit being a nearly perfect diagonal line. FLAML also leans towards specific models based on the data and will iterate more on more suitable models, for example, for Dataset 2, FLAML tends to always select Light GBM as the model used and for Dataset 3, FLAML tends to avoid models like Random Forest. Within this work, we use FLAML in order to train accurate and complicated models, in order to show the utility of LIME and to ensure that LIME is truly being tested, not the abilities of FLAML.

\begin{figure}[htp]
    \centering
    
    \begin{subfigure}[b]{0.23\textwidth}
        \includegraphics[width=\textwidth]{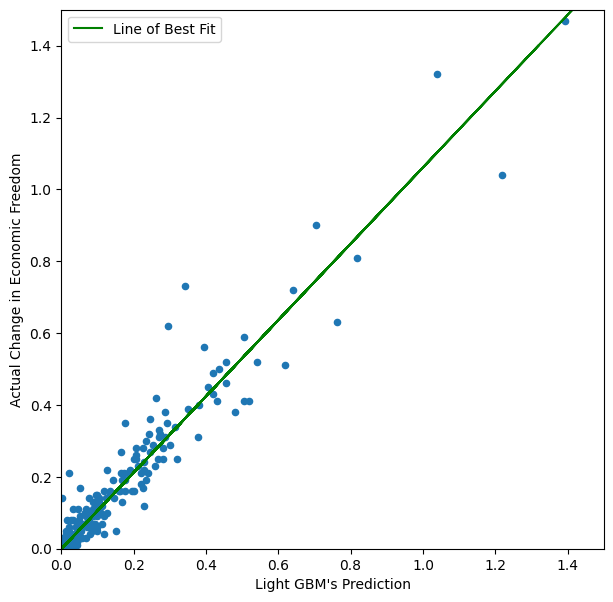}
        \caption{Predictions by Light GBM for Dataset 2}
        \label{fig:lightgbm}
    \end{subfigure}
    \hfill
    \begin{subfigure}[b]{0.23\textwidth}
        \includegraphics[width=\textwidth]{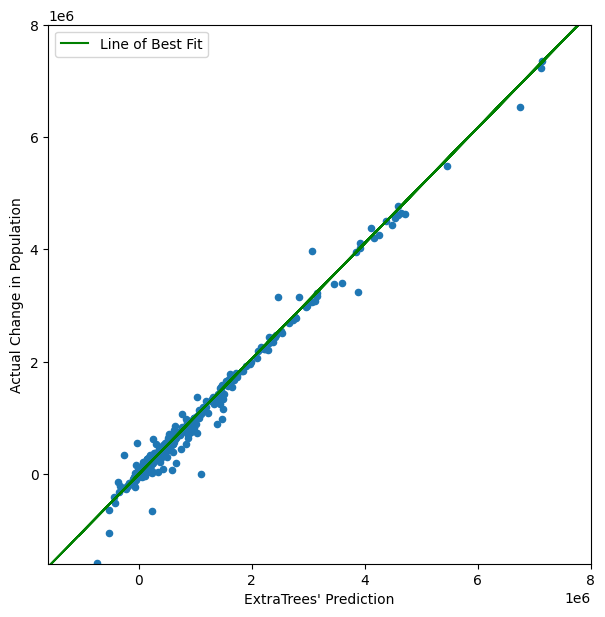}
        \caption{Predictions by ExtraTrees for Dataset 3}
        \label{fig:extratree}
    \end{subfigure}

    \caption{Predictions by different models for Dataset 2 and Dataset 3}
    \label{fig:combined_figure}
\end{figure}

\subsection{Interpreting with LIME}

While complex models, such as deep neural networks or ensemble methods generated from FLAML, offer high accuracy, they often lack transparency, making it challenging to understand their decision-making process. LIME addresses this by locally approximating the black-box model's predictions. For a given prediction, LIME generates a set of perturbed data samples, obtains predictions for these samples from the black-box model, and then fits a simpler, interpretable model to these samples, typically a linear regression model. LIME figures out what the most effective columns are through these perturbations \cite{b18}. For example, if changing the value of the country's area greatly affects the prediction for the country's population, then this might be an important feature. Consequently, if changing a feature affects the result by a minor amount or not at all, then the feature isn't important.

For this purpose, we set $kernel\_width$ to $1, 0.849$, and $9000$ for three datasets respectively, where the kernel width determines the width of the neighborhood of points around the data point that's allowed to influence the local model and consequently, the explanation. The models built by LIME serve as a proxy to the black-box model but is transparent in its decision-making. In order to properly test the quality of LIME, we ran comprehensive tests, where LIME made predictions for the entire test set in the case of the first two datasets and for the first 250 rows in the case of the third test set, while being restricted to a set number of columns, with that number being 3 for the first dataset and 10 for the other two datasets. The remainder of the rows' values were set to 0. The R\textsuperscript{2} values were collected and were compared to cases where the same number of columns were randomly selected and the original model's R\textsuperscript{2} values. Therefore, we can show that LIME's selection of columns is significantly better than randomly selecting columns, as the brilliance of LIME lies in its ability to provide insight into the black-box model's predictions for individual data points rather than a global approximation.

To further justify the results we receive from LIME, we compare its results with those derived from Individual Conditional Expectation (ICE) plots \cite{b19}, which are a tool used for visualizing the model predictions for individual instances as a function of specific input features. ICE plots are an extension of Partial Dependence Plots (PDPs) \cite{b20}, which show the average prediction of a machine learning model as a function of one or two input features, holding other features constant. As ICE plots offer a nuanced view of the relationship between specific features and the predicted outcome, features with ICE lines that show significant slopes suggest that a small change in the feature value can lead to a substantial change in the prediction. A steep slope often signifies feature importance, thus, as LIME assigns importance scores to features, by comparing the top columns returned by LIME with the slopes and variations observed in ICE plots and checking if frequently selected columns have steep slopes, we can have improved confidence in whether LIME is correctly identifying important features.

\section{Results and Discussions}

\subsection{Experiment Setting}

For each run of the FLAML algorithm, a time span of $T=180$ seconds is allocated with the aim to optimize the \( R^2 \) value in our study. The hyperparameters that are selected vary with each run, as well.
The \( R^2 \) value, commonly known as the coefficient of determination, quantifies the proportion of the variance in the dependent variable that can be predicted from the independent variables. Essentially, it gauges the extent to which the independent variables in a model elucidate the variability of the dependent variable. 
The formula for \( R^2 \) is given by:
\begin{equation}
R^2 = \frac{\sum_{i=1}^{n}(y_i-\hat{y_i})^2}{\sum_{i=1}^{n}(y_i-\bar{y})^2}
\end{equation}
In the above equation:
\begin{itemize}
    \item \( y_i \) denotes the observed value.
    \item \( \hat{y_i} \) symbolizes the value predicted by the model.
    \item \( \bar{y} \) represents the mean of the observed values.
\end{itemize}
The numerator encapsulates the squared discrepancies between the actual observations and the predictions made by the model, reflecting the variance that the model fails to explain. Meanwhile, the denominator computes the total variance present in the data.
The possible range for \( R^2 \) is between 0 and 1:
\begin{itemize}
    \item An \( R^2 \) value of 0 suggests that the model fails to explain any variability of the dependent variable around its average.
    \item An \( R^2 \) value of 1 indicates that the model accounts for all the variability.
\end{itemize}
For Dataset 1, the models considered include Light GBM, Random Forest, XGBoost, ExtraTrees, and a depth-limited XGBoost. In the case of Dataset 2, only Light GBM is utilized, while for Dataset 3, all models except Random Forest are employed. These are all variants that support regression. Model selection is refined based on frequency: models that are either consistently favored or rarely chosen are omitted, allowing the algorithm to focus on the more influential models. Initially, long short-term memory networks (LSTMs) are explored, but are later excluded due to their unsatisfactory \( R^2 \) preliminary results, even when using simpler LSTM configurations. A potential rationale for this exclusion is that all three datasets have a relatively small size, which might hinder the LSTM's ability to construct a sturdy model. The algorithm was run five times for each dataset, with the results being provided in Table \ref{table:I}. The mean \( R^2 \) values are included to give the viewer an overall impression of the results.

\begin{table}[ht]
\begin{center}
\begin{tabular}{|c c c c|} 
 \hline
 Statsitic & Dataset 1 & Dataset 2 & Dataset 3 \\ [0.5ex] 
 \hline\hline
 Run \#1 & 0.092 & 0.924 & 0.988\\ 
 \hline
 Run \#2 & 0.036 & 0.868 & 0.995\\
 \hline
 Run \#3 & 0.21 & 0.846 & 0.994\\
 \hline
 Run \#4 & 0.071 & 0.904 & 0.992\\
 \hline
 Run \#5 & 0.226 & 0.883 & 0.996\\
 \hline
 Mean & 0.127 & 0.885 & 0.993\\
 \hline
\end{tabular}
\caption{A table showing the $R^2$ values for each of the datasets with FLAML.}
\label{table:I}
\end{center}
\end{table}

\subsection{LIME's Results}


To assess the effectiveness of LIME in feature selection, we carried out multiple experiments, contrasting random column selection with LIME-guided column selection. For each of the three datasets, these experiments were conducted five times to mitigate the impact of randomness, with the average results presented in Fig. 2. Remarkably, across all three sets of experiments, LIME-guided feature selection consistently outperformed random selection in terms of \( R\textsuperscript{2} \) values. This was true even for Dataset 1, which is the most difficult to understand. Moreover, there was an absolute increase in the \( R\textsuperscript{2} \) values in every instance. This was particularly noteworthy for the third dataset, where the \( R\textsuperscript{2} \) values surged by an average of 0.4904 when employing LIME for feature selection.

\begin{figure}[htp]
    \centering
    \includegraphics[width=10cm]{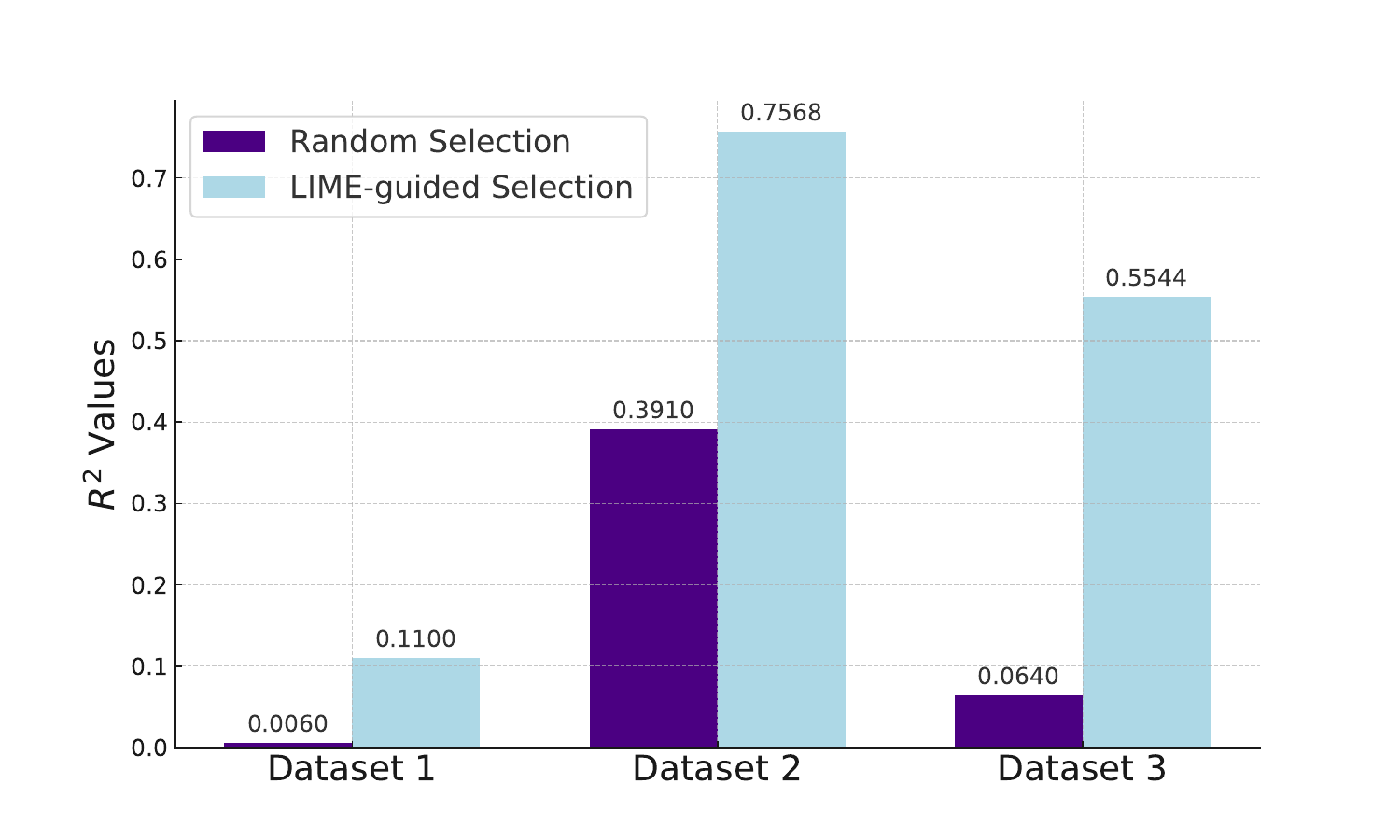}
    \caption{A bar graph comparing the $R^2$ values of selecting random columns VS selecting columns with LIME.}
\end{figure}

This can be further verified with an one-sided Paired T-Test, which compares the results without the use of LIME to the results with LIME. This test showed significance for the first dataset at $\alpha$ = 0.02, while the latter two results showed significance at $\alpha$ = 0.005, therefore, the approach that was taken by LIME was clearly statistically significant and beneficial compared to random predictions. Most notably, the P-value for Dataset 2 was 0.00005922. Dataset 1 was difficult to interpret and make predictions for, but we've shown that some data and insights could be discovered, hence why the R\textsuperscript{2} were significant at $\alpha$ = 0.02, while Datasets 2 and 3 could be fully interpreted by LIME and returned excellent results for the R\textsuperscript{2} values. LIME was equally beneficial for individual predictions, as will be demonstrated through the use of real-world scenarios and ICE Plots.

\subsection{Case Studies}



To illustrate LIME's efficacy, we opted for a case study approach using two extreme cases from Dataset 2. This approach not only validates the quality of LIME but also lends real-world context to its utility. The key idea is to match significant changes in a country's economic freedom with major historical or socio-political events and see if LIME's explanations align with these changes.\\
For instance, Syria experienced a substantial decline in its economic freedom during 2011 and 2012, largely due to the onset of the Syrian civil war, a part of the broader Arab Spring movements\footnote{\url{https://www.atlanticcouncil.org/blogs/menasource/the-economic-collapse-of-syria/}}. In 2011, Syria's actual economic freedom score plummeted by 0.91 points. When subjected to LIME's analysis, the algorithm estimated a significant drop and associated this decline primarily with three factors: inflation, credit market regulation, and regulatory trade barriers. Their respective scores in the dataset decreased by 6.39, 1.4, and 0.56 points between 2011 and 2012, which align with the observations found in Fig 3. It's worth noting that Syria faced an inflation rate of 36.7\% \footnote{\url{https://www.macrotrends.net/countries/SYR/syrian-arab-republic/inflation-rate-cpi}} in 2012 and also experienced severe disruptions in its business environment, leading to a mass exodus of foreign investors and tourists. These real-world factors align well with LIME's explanation, further validating its effectiveness.

\begin{figure}[htp]
    \centering
    \includegraphics[width=9cm]{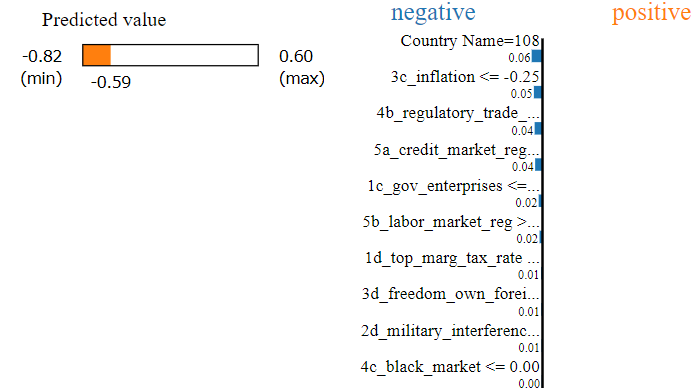}
    \caption{A prediction made by LIME showing the columns that most contributed to Syria's drop in Economic Freedom between 2011 and 2012.}
\end{figure}

Inversely, Brazil had one of the largest improvements in the dataset, with a strong comeback that took place from 1995 to 2000 and raised the country's economic freedom by 1.32 points. This was mainly in part due to previous concerns that the country had in 1990 to 1995 being fixed. For example, Brazil suffered from massive inflation, first in early 1990 and then in 1994, with yearly rates of 2,947.73\% and 2,075.89\%, respectively, however, after the Plano Real, a stabilization program, was introduced to curb this, the largest yearly inflation from 1996-2000 was merely 15.76\%, in 1996 \footnote{\url{https://www.macrotrends.net/countries/BRA/brazil/inflation-rate-cpi}}. A black market also existed, due to the ineffectiveness of actual money and the need for affordable food and water. Finally, money growth also greatly increased, which is calculated through the average annual growth of a country's money supply over the last five years minus average annual growth of real GDP in the last ten years. Annual money growth was massive in 1994 and at over 6,000\% in 1994 \footnote{\url{https://www.ceicdata.com/en/indicator/brazil/m2-growth}}, but would never reach such heights again and remained below 30\% during the five-year period, largely in part because of the disastrous inflation being fixed.

\begin{figure}[htp]
    \centering
    \includegraphics[width=9cm]{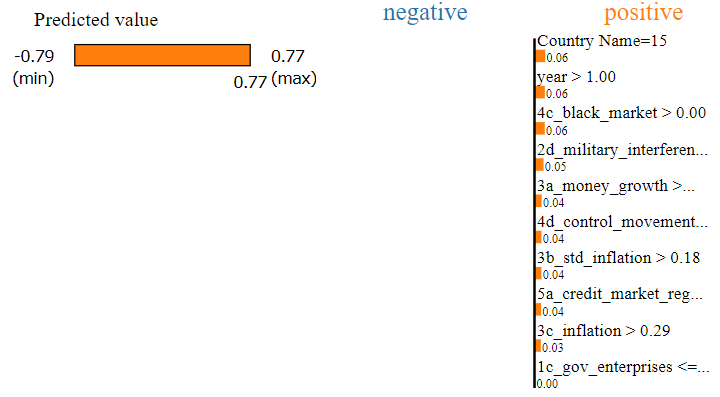}
    \caption{A prediction made by LIME showing the columns that most contributed to the rise in Brazil's Economic Freedom from 1995 to 2000.}
\end{figure}

\subsection{Comparison with State-of-the-art Methods}

In order to further justify the effectiveness of LIME, we used Individual Conditional Explanation plots, acting as an extension of Partial Dependence Plots, to allow for a further analysis of LIME by analyzing the slopes of each feature. A feature with a steeper slope than other features in the ICE plot will often indicate feature importance. Thus, by comparing the columns provided by LIME with the ICE plots, we can assess the quality of LIME's predictions and show that the ICE plots verify our results.

In our study, ICE was employed to scrutinize changes at the level of individual instances. To support this, as an example 20 rows were chosen using LIME's Submodular Pick function for Dataset 3, which assembles a series of local, minimally redundant explanations. The selection of 20 rows aimed to offer comprehensive coverage, align coherently with ICE plots that also focus on individual rows, and yield conclusive insights. We generated Table \ref{table:4} detailing the frequency with which specific columns appeared among the top 5 chosen for these explanations. Limiting the selection to 5 columns, rather than 10, helps to concentrate attention on the most influential features. Our hypothesis posits that the features selected by LIME—due to its focus on perturbing features to identify those with the most significant impact—will corroborate their importance when examined through ICE plots, thus making them ideal candidates for generating explanations.

\begin{table}[h!]
\begin{center}
\begin{tabular}{|c c|} 
 \hline
 Feature Name & Frequency\\ [0.5ex] 
 \hline\hline
 Area in Square Kilometers & 20\\ 
 \hline
 Sex Ratio at Birth & 17\\
 \hline
 Infant Mortality Rate for Females & 14\\
 \hline
 Under Age 5 Mortality for Females & 12\\
 \hline
 Age 1-4 Mortality for Females & 9\\
 \hline
 Age-Specific Fertility Rate 20-24 & 7\\
 \hline
 Infant Mortality Rate for Both Sexes & 4\\
 \hline
 Age 1-4 Mortality for Both Sexes & 3\\
 \hline
 Crude Death Rate & 3\\
 \hline
 Infant Mortality Rate for Males & 2\\
 \hline
 Age-Specific Fertility Rate 25-29 & 2\\
 \hline
 Age-Specific Fertility Rate 35-39 & 2\\
 \hline
 Under Age 5 Mortality for Males & 1\\
 \hline
 Under Age 5 Mortality for Both Sexes & 1\\
 \hline
 Age-Specific Fertility Rate 40-44 & 1\\
 \hline
 Life Expectancy at Birth for Males & 1\\
 \hline
 Life Expectancy at Birth for Females & 1\\
 \hline
\end{tabular}
\caption{A frequency table of how LIME selects features in Dataset 3.}
\label{table:4}
\end{center}
\end{table}
From Table \ref{table:4}, it is obvious that even with the algorithm trying to make the explanations as varied as possible, the feature Area in Square Kilometers was selected every single time, while Sex Ratio at Birth, Infant Mortality Rate for Females and Under Age 5 Mortality for Females were also commonly selected. 
\begin{figure}[htp]
    \centering
    \includegraphics[width=8.5cm]{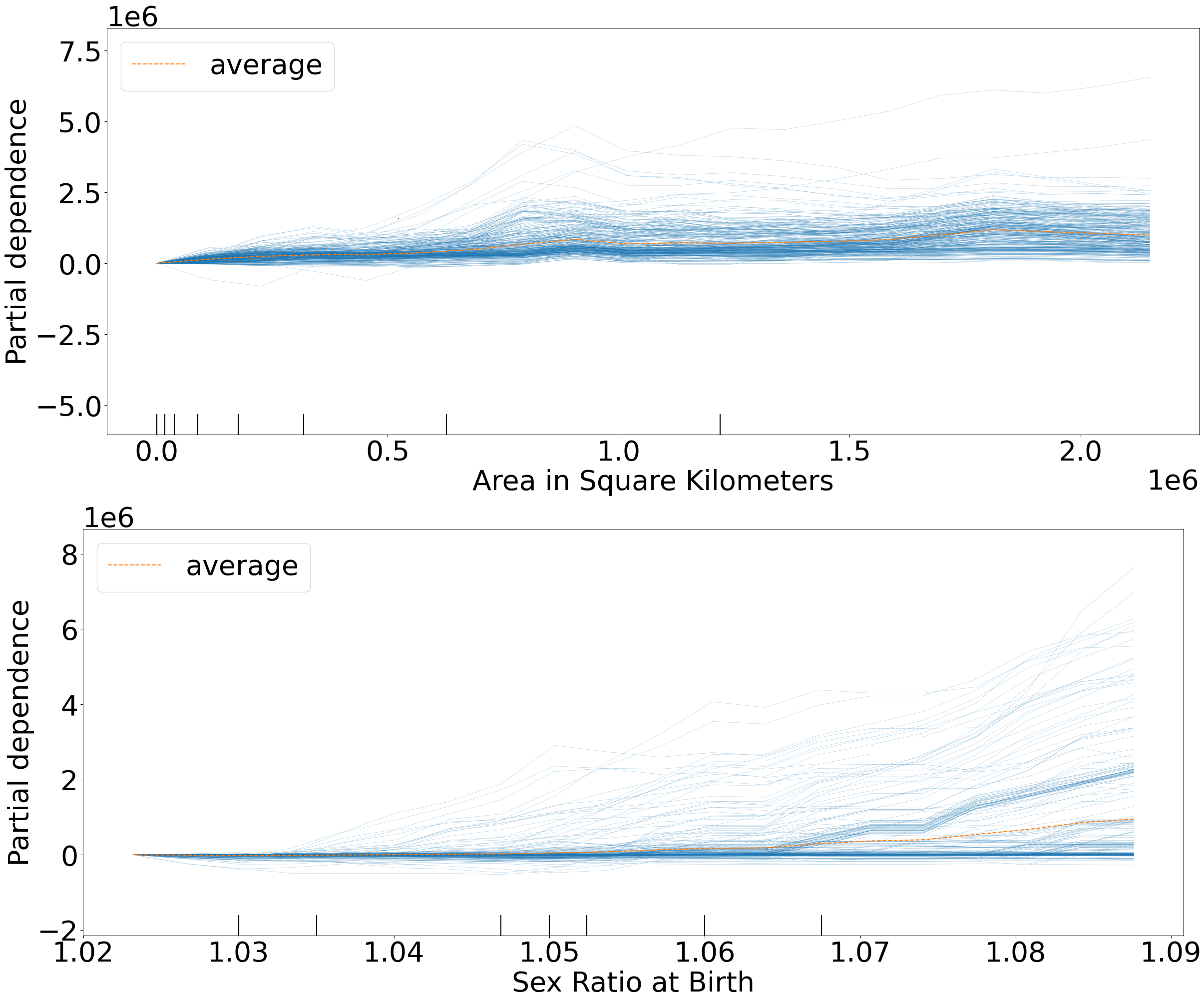}
    \caption{ICE Plots showing how greatly the columns that are most selected by LIME affect the change in population, for Dataset 3.}
    \label{fig:ice}
\end{figure}

It's clear that these selected columns are backed up by the ICE Plots in Fig. \ref{fig:ice}, which all have large variances and can affect the change in a country's population by several hundred thousand. We also noted that the features Area in Square Kilometers and Sex Ratio at Birth have the steepest slopes. In conclusion, LIME and ICEs both agree that these columns were the most important to Dataset 3's predictions.

\section{Conclusion}


In summary, this research presents a rigorous exploration into the use of LIME for multivariate temporal analysis on global annual datasets. Our approach successfully identifies key factors contributing to temporal shifts, offering explanations that are critical for both academic inquiry and practical applications. Through the integration of LIME with other advanced methods such as ICE plots, we have not only validated the efficacy of our approach but also opened new avenues for explainable AI in temporal analysis.

The implementation of multiple imputation techniques to handle missing data in the second dataset exemplifies the adaptability and robustness of our method. Approximately 65\% of the records in this dataset had missing values, and our strategy ensured reliable, high-quality imputations that enriched the overall analysis.

Furthermore, our study lays the groundwork for future research in this area. The analytical frameworks presented here can be extended to tackle more complex datasets and questions, offering transformative impacts that range from the reshaping of governmental policies to steering economic strategies.

Future research could focus on comparing the effectiveness of LIME against other explainability models or delve into real-world applications that could benefit from our approach.

\bibliographystyle{ieeetran}
\bibliography{lime}

\begin{thebibliography}{10}
\providecommand{\url}[1]{#1}
\csname url@samestyle\endcsname
\providecommand{\newblock}{\relax}
\providecommand{\bibinfo}[2]{#2}
\providecommand{\BIBentrySTDinterwordspacing}{\spaceskip=0pt\relax}
\providecommand{\BIBentryALTinterwordstretchfactor}{4}
\providecommand{\BIBentryALTinterwordspacing}{\spaceskip=\fontdimen2\font plus
\BIBentryALTinterwordstretchfactor\fontdimen3\font minus
  \fontdimen4\font\relax}
\providecommand{\BIBforeignlanguage}[2]{{%
\expandafter\ifx\csname l@#1\endcsname\relax
\typeout{** WARNING: IEEEtran.bst: No hyphenation pattern has been}%
\typeout{** loaded for the language `#1'. Using the pattern for}%
\typeout{** the default language instead.}%
\else
\language=\csname l@#1\endcsname
\fi
#2}}
\providecommand{\BIBdecl}{\relax}
\BIBdecl

\bibitem{b3}
M.~T. Ribeiro, S.~Singh, and C.~Guestrin, ``"why should i trust you?":
  Explaining the predictions of any classifier,'' in \emph{Proceedings of the
  22nd ACM SIGKDD International Conference on Knowledge Discovery and Data
  Mining}, 2016, p. 1135–1144.

\bibitem{b17}
A.~Goldstein, A.~Kapelner, J.~Bleich, and E.~Pitkin, ``Peeking inside the black
  box: Visualizing statistical learning with plots of individual conditional
  expectation,'' \emph{Journal of Computational and Graphical Statistics},
  vol.~24, pp. 44 -- 65, 2013.

\bibitem{b1}
M.~Guillemé, V.~Masson, L.~Rozé, and A.~Termier, ``Agnostic local explanation
  for time series classification,'' in \emph{2019 IEEE 31st International
  Conference on Tools with Artificial Intelligence (ICTAI)}, 2019, pp.
  432--439.

\bibitem{b2}
T.~A. Abdullah, M.~S. B.~M. Zahid, T.~B. Tang, W.~Ali, and M.~Nasser,
  ``Explainable deep learning model for cardiac arrhythmia classification,'' in
  \emph{2022 International Conference on Future Trends in Smart Communities
  (ICFTSC)}, 2022, pp. 87--92.

\bibitem{b13}
C.~Wang, Q.~Wu, M.~Weimer, and E.~Zhu, ``Flaml: A fast and lightweight automl
  library,'' in \emph{Conference on Machine Learning and Systems}, 2019.

\bibitem{b4}
U.~Schlegel, D.~L. Vo, D.~A. Keim, and D.~Seebacher, ``Ts-mule: Local
  interpretable model-agnostic explanations for time series forecast models,''
  in \emph{Machine Learning and Principles and Practice of Knowledge Discovery
  in Databases}, 2021, pp. 5--14.

\bibitem{b5}
T.~Sivill and P.~Flach, ``Limesegment: Meaningful, realistic time series
  explanations,'' in \emph{Proceedings of The 25th International Conference on
  Artificial Intelligence and Statistics}, ser. Proceedings of Machine Learning
  Research, G.~Camps-Valls, F.~J.~R. Ruiz, and I.~Valera, Eds., vol. 151,
  28--30 Mar 2022, pp. 3418--3433.

\bibitem{b6}
M.~Nikam, A.~Ranade, R.~Patel, P.~Dalvi, and A.~Karande, ``Explainable approach
  for species identification using lime,'' in \emph{2022 IEEE Bombay Section
  Signature Conference (IBSSC)}, 2022, pp. 1--6.

\bibitem{b7}
S.~Sahay, N.~Omare, and K.~K. Shukla, ``An approach to identify captioning
  keywords in an image using lime,'' in \emph{2021 International Conference on
  Computing, Communication, and Intelligent Systems (ICCCIS)}, 2021, pp.
  648--651.

\bibitem{b8}
S.~Rao, S.~Mehta, S.~Kulkarni, H.~Dalvi, N.~Katre, and M.~Narvekar, ``A study
  of lime and shap model explainers for autonomous disease predictions,'' in
  \emph{2022 IEEE Bombay Section Signature Conference (IBSSC)}, 2022, pp. 1--6.

\bibitem{b9}
A.~Torcianti and S.~Matzka, ``Explainable artificial intelligence for
  predictive maintenance applications using a local surrogate model,'' in
  \emph{2021 4th International Conference on Artificial Intelligence for
  Industries (AI4I)}, 2021, pp. 86--88.

\bibitem{b10}
J.~F. Helliwell, R.~Layard, J.~D. Sachs, L.~B. Aknin, J.-E. De~Neve, and
  S.~Wang, Eds., \emph{World Happiness Report 2023}, 11st~ed.\hskip 1em plus
  0.5em minus 0.4em\relax Sustainable Development Solutions Network, 2023.

\bibitem{b11}
G.~S. Schneider. (2018) Economic freedom of the world. Accessed on 19th October
  2018.

\bibitem{b12}
{U.S. Census Bureau}. (2020) International database (idb). Accessed in December
  2020.

\bibitem{b18}
C.~Molnar, \emph{Interpretable machine learning}.\hskip 1em plus 0.5em minus
  0.4em\relax Lulu. com, 2020.

\bibitem{b19}
A.~Goldstein, A.~Kapelner, J.~Bleich, and E.~Pitkin, ``Peeking inside the black
  box: Visualizing statistical learning with plots of individual conditional
  expectation,'' 2014.

\bibitem{b20}
J.~H. Friedman, ``{Multivariate Adaptive Regression Splines},'' \emph{The
  Annals of Statistics}, vol.~19, no.~1, pp. 1 -- 67, 1991.

\end{thebibliography}

\onecolumn

\appendix
\section{Dataset Snippets}\label{sec:app}

The following tables display some randomly selected records from each dataset, to provide an understanding of the challenges faced with the missing data in Dataset 2 and an idea of what kind of data was analyzed within the paper. 

\begin{table}[h!]
\begin{center}
\begin{tabular}{|c c c c c c c c|} 
 \hline
 Country Name & Year & Life Ladder & Log GDP per Capita & Social Support & Life Expectancy & ... & Dystopia + Residual\\ [0.5ex] 
 \hline
 Gabon & 2021 & 5.075 & 9.533 & 0.754 & 58.25 &  ... & 2.201\\ 
 \hline
 Honduras & 2016 & 5.648 & 8.573 & 0.774 & 61.725 &  ... & 2.296\\
 \hline
 Poland & 2020 & 6.139 & 10.39 & 0.953 & 68.875 &  ... & 2.056798\\
 \hline
 Togo & 2022 & 4.239 & 7.685 & 0.579 & 57.7 &  ... & 2.061\\
 \hline
 United States & 2016 & 6.804 & 10.985 & 0.897 & 66.475 &  ... & 2.728\\
 \hline
 Zambia & 2018 & 4.041 & 8.139 & 0.718 & 53.975 &  ... & 1.667\\
 \hline
\end{tabular}
\caption{A snippet of Dataset 1.}
\label{table:5}
\end{center}
\end{table}

\begin{table}[h!]
\begin{center}
\begin{tabular}{|c c c c c c c|} 
 \hline
 Year & Country & Economic Freedom & ... & 5a\_credit\_market\_reg & 5b\_labor\_market\_reg & 5c\_business\_reg\\ [0.5ex] 
 \hline
 2012 & Australia & 7.96 & ... & 9.596215 & 6.744523 & 8.036912\\ 
 \hline
 1995 & Bangladesh & 5.3 & ... & 5.122456 & N/A & N/A\\
 \hline
 2003 & Benin & 6.03 & ... & 9.167872 & 4.665644 & N/A\\
 \hline
 1990 & Switzerland & 8.21 & ... & 8.153153 & 5.4835 & N/A\\
 \hline
 2012 & Tanzania & 6.62 & ... & 8.985725 & 6.477967 & 5.588262\\
 \hline
 1995 & Uganda & 5.09 & ... & 4.285619 & N/A & N/A\\
 \hline
 2011 & Syria & 6.18 & ... & 6.981081 & 4.841801 & 5.970616\\
 \hline
 2012 & Syria & 5.27 & ... & 5.585419 & 5.432596 & 4.987837\\
 \hline
 1995 & Brazil & 4.66 & ... & 4.573693 & 5.296847 & 4.942583\\
 \hline
 2000 & Brazil & 5.98 & ... & 6.228699 & 4.82906 & 5.470623\\ 
 \hline
\end{tabular}
\caption{A snippet of Dataset 2.}
\label{table:6}
\end{center}
\end{table}

\begin{table}[h!]
\begin{center}
\begin{tabular}{|c c c c c c c|} 
 \hline
 Country/Area Name & Year & Area in KM\textsuperscript{2} & Population & ... & Crude Death Rate & Net Migration Rate\\ [0.5ex] 
 \hline
 Andorra & 2010 & 468 & 84,563 & ... & 5.89 & 0\\ 
 \hline
 Cambodia & 2022 & 176,515 & 16,713,015 & ... & 5.76 & -2.7\\
 \hline
 Indonesia & 2023 & 1,811,569 & 279,476,346 & ... & 6.77 & -0.71\\
 \hline
 Mauritania & 2012 & 1,030,700 & 3,356,215 & ... & 8.77 & -0.89\\
 \hline
 Spain & 2005 & 498,980 & 43,704,367 & ... & 8.86 & 14.55\\
 \hline
 United States & 2023 & 9,150,541 & 339,665,118 & ... & 8.4231 & 3.0116\\
 \hline
\end{tabular}
\caption{A snippet of Dataset 3.}
\label{table:7}
\end{center}
\end{table}

\end{document}